\documentclass[runningheads]{llncs}
\usepackage{enumitem}
\usepackage{graphicx}
\usepackage{amsmath}
\usepackage{amssymb}
\usepackage{booktabs}
\usepackage{color}
\usepackage{alltt}
\usepackage{amsfonts}
\usepackage{mathrsfs}
\usepackage{algorithm}
\usepackage{comment}
\usepackage{hyperref}
\usepackage{marvosym}
\usepackage{algorithmic}
\usepackage{amssymb}
\usepackage{pifont}
\usepackage{wrapfig}
\usepackage{fancyvrb} 
\hypersetup{hidelinks}
\usepackage{multicol,multirow}
\usepackage[dvipsnames]{colortbl}
\usepackage[accsupp]{axessibility}
\newcommand{\cmark}{\ding{51}}%
\newcommand{\xmark}{\ding{55}}%

\begin{document}
\pagestyle{headings}
\mainmatter

\title{Bootstrap Generalization Ability from Loss Landscape Perspective}
\def\ECCVSubNumber{12} 

\titlerunning{Bootstrap Generalization Ability from Loss Landscape Perspective}
%
\author{Huanran Chen\inst{1}
\and
Shitong Shao\inst{2}
\and
Ziyi Wang\inst{1}
\and
Zirui Shang\inst{1}
\and
Jin Chen\inst{1}
\and
Xiaofeng Ji\inst{1}
\and
Xinxiao Wu\inst{1}\textsuperscript{\href{mailto:wuxinxiao@bit.edu.cn}{\Letter}
}}
\authorrunning{Chen et al.}
\institute{Beijing Laboratory of Intelligent Information Technology,\\ 
School of Computer Science, Beijing Institute of Technology, Beijing, China \email{\{huanranchen,wangziyi,shangzirui,chen\_jin,jixf,wuxinxiao\}@bit.edu.cn}\and
Southeast University, Nanjing, China\\
\email{shaoshitong@seu.edu.cn}}
\maketitle

\begin{abstract}
Domain generalization aims to learn a model that can generalize well on the unseen test dataset, \emph{i.e.,} out-of-distribution data, which has different distribution from the training dataset. To address domain generalization in computer vision, we introduce the loss landscape theory into this field. Specifically, we bootstrap the generalization ability of the deep learning model from the loss landscape perspective in four aspects, including backbone, regularization, training paradigm, and learning rate. We verify the proposed theory on the NICO++, PACS, and VLCS datasets by doing extensive ablation studies as well as visualizations. In addition, we apply this theory in the ECCV 2022 NICO Challenge1 and achieve the 3rd place without using any domain invariant methods.

\end{abstract}

\section{Introduction}
Deep learning has made great progress in many popular models (\emph{e.g.}, ViT ~\cite{dosovitskiy2020image}, Swin-T~\cite{SWIN-T}, GPT3~\cite{brown2020language}) when the training data and test data satisfy the identically and independently distributed assumption. However, in real applications, this assumption may be hard to hold due to the considerable variances in object appearances, illumination, image style, \emph{etc.}~\cite{ben2010theory}. Such variances between training data (source domain) and test data (target domain) are denoted as the domain shift problem and have been addressed via Domain Adaptation (DA) and Domain Generalization (DG)~\cite{wang2022generalizing}. DA requires the access of target domains. In contrast, DG aims to learn a model with good generalization from multiple source domains without the requirement of target domains, receiving increasing scholarly attention. Existing DG methods can be divided into three categories: data augmentation~\cite{wang2021learning,shankar2018generalizing}, domain-invariant representation learning~\cite{ajakan2014domain,pan2010domain}, and learning strategy optimization~\cite{huang2020self,balaji2018metareg}.

To analyze the domain generalization problem comprehensively, we have tried the popular methods of the three categories in many datasets. According to experiment results, we have several interesting observations. First, domain-invariant representation learning methods (DANN~\cite{mao2017least} and MMD~\cite{long2015learning}) perform worse or even hurt the performance on unseen test domains if there is no significant gap between each domain in the training set and test set. The reason may be that the domain shift across source domains are mainly caused by the small variances of context, making the distribution across source domains similar. Under this situation, brute-force domain alignment makes the model over-fit source domains and generalize bad on unseen target domains. Second, data augmentation and learning strategy optimization can substantially improve the model's generalization ability on this kind of scenarios. A method of combining data augmentation-based approaches and careful hyper-parameter tuning has achieved state-of-the-art performance~\cite{gulrajani2020search}. We found that most of these methods can be explained from the perspective of landscape. By increasing the training data and optimizing the learning strategy, the model can be trained to converge to a flat optimum, and thus achieves good result in the test set, which inspires us to use landscape theory to give explanations of phenomena and methods in domain generalization and to devise more useful methods.

It has been both empirically and theoretically shown that a flat minimum leads to better generalization performance. A series of methods such as SAM~\cite{foret2020sharpness}, delta-SAM\cite{zhou2021delta}, and ASAM~\cite{kwon2021asam} have been proposed to allow the model to converge to a more flat minimum. Inspired by this, we propose a series of learning strategies based on loss landscape theory that enable the model to perform well on the NICO++ dataset. First, we embed supervised self-distillation into the fine-tuning paradigm by interpreting it as obtaining a flatter loss landscape, ultimately proposing a novel Distillation-based Fine-tuning Paradigm (DFP). Furthermore, we observe that the convergence to the minimum point is relatively flat when the learning rate is large. We argue that effective learning rate control can directly make the model converge to the extreme flat point. We propose a new learning rate scheduler, called Adaptive Learning Rate Scheduler (ALRS), which can automatically schedule the learning rate and let the model converge to a more flat minimum. Finally, according to the theory of~\cite{han2017deep} and~\cite{2015The}, wider models are more easily interpreted as convex minimization problems. Therefore, we widen the traditional PyramidNet272 and name it Wide PyramidNet272, which can be interpreted in loss landscape theory to obtain a smoother surface and a flatter convergence region. By simultaneously applying all the methods we have proposed, our method achieves superior performance against the state-of-the-art methods on public benchmark datasets, including NICO++, PACS and VLCS.

In general, our contributions can be summarized as follows:
\begin{itemize}[topsep=-5pt]
\setlength{\itemsep}{0pt}
\setlength{\parsep}{0pt}
\setlength{\parskip}{0pt}
        \item[$\bullet$] We explain some of the existing methods in DG from the perspective of loss landscape, such as ShakeDrop~\cite{yamada2019ShakeDrop}, Supervised Self-Distillation (MESA)~\cite{SAM}.
        \item[$\bullet$] Inspired by loss landscape theory, we propose some general methods, such as DFP, ALRS, and Super Wide PyramidNet272.
        \item[$\bullet$] We have done extensive experiments on NICO++, PACS, and VLCS datasets and visualizations to justify these explanations.
\end{itemize}
\section{Related Work}
\paragraph{Scheduler.} Learning rate is an essential hyper-parameter in training neural networks. Lots of research has shown that different learning rates are required at different stages of training~\cite{Zagoruyko2016WRN,ResNet,cosinewarmup}. Commonly, a lower learning rate is required in the early (w.r.t., warmup) and late stages of training, and a larger learning rate is required in the middle of training to save time. Researchers have designed lots of methods to achieve this goal, such as Linear Warm-up~\cite{goyal2017accurate}, Cosine Annealing~\cite{DBLP:journals/corr/LoshchilovH16a}, Exponential Decay, Polynomial Rate Decay, Step Decay, 1cycle~\cite{smith2018disciplined}. However, they all assume that the learning rate is a function of the training epoch rather than the model performance. To fill this gap, we propose a new learning rate scheduler, which can automatically adjust the learning rate according to the performance of the model.
\paragraph{Fine-tuning.} Fine-tuning is also an important technique that allows the model to quickly adapt to the target dataset in a shorter time with fewer data. In addition to this, it allows the model to adapt to large-resolution inputs quickly. To be specific, pretraining with small resolution and fine-tuning with large resolution on the target dataset, can effectively reduce computational costs and ensure generalization performance of the model. Therefore, we incorporate MESA~\cite{SAM}, a method similar to self-distillation, into the fine-tuning paradigm and propose a distillation-based fine-tuning paradigm. We will discuss this in more details in Section 3.1.
\paragraph{Data Augmentation.} Data augmentation is an important strategy to expand the training dataset. It translates, rotates, scales, and mixes images to generate more equivalent data that can be learned~\cite{PBA,Mixup,cutmix}. Recent studies have shown that the optimal data augmentation strategies manifest in various ways under different scenarios. AutoAugment~\cite{cubuk2018autoaugment} and RandAugment~\cite{cubuk2020randaugment}, two popular automatic augmentations, are experimentally validated for their ability to improve model accuracy on many real datasets. Since we do not have enough resources to search for optimal hyper-parameters, we adopt the AutoAugment strategy designed for CIFAR-10. As demonstrated experimentally in this paper, even just importing the AutoAugment strategy on CIFAR-10, can significantly improve the model's performance.
\paragraph{Loss Landscape.} It has been both empirically and theoretically shown that a flat minimum leads to better generalization performance. Zhang et al.~\cite{zhang2021flatness} empirically and theoretically show that optima's flatness correlates with the neural network's generalization ability. A series of methods based on SAM has been developed to allow the model to converge to a flat optimum~\cite{SAM,du2021efficient,kwon2021asam}. As a result, we present a series of methods based on the loss landscape theory to boost the generalization ability of our model. 

\begin{figure}[t]
\centering
\includegraphics[width=\textwidth]{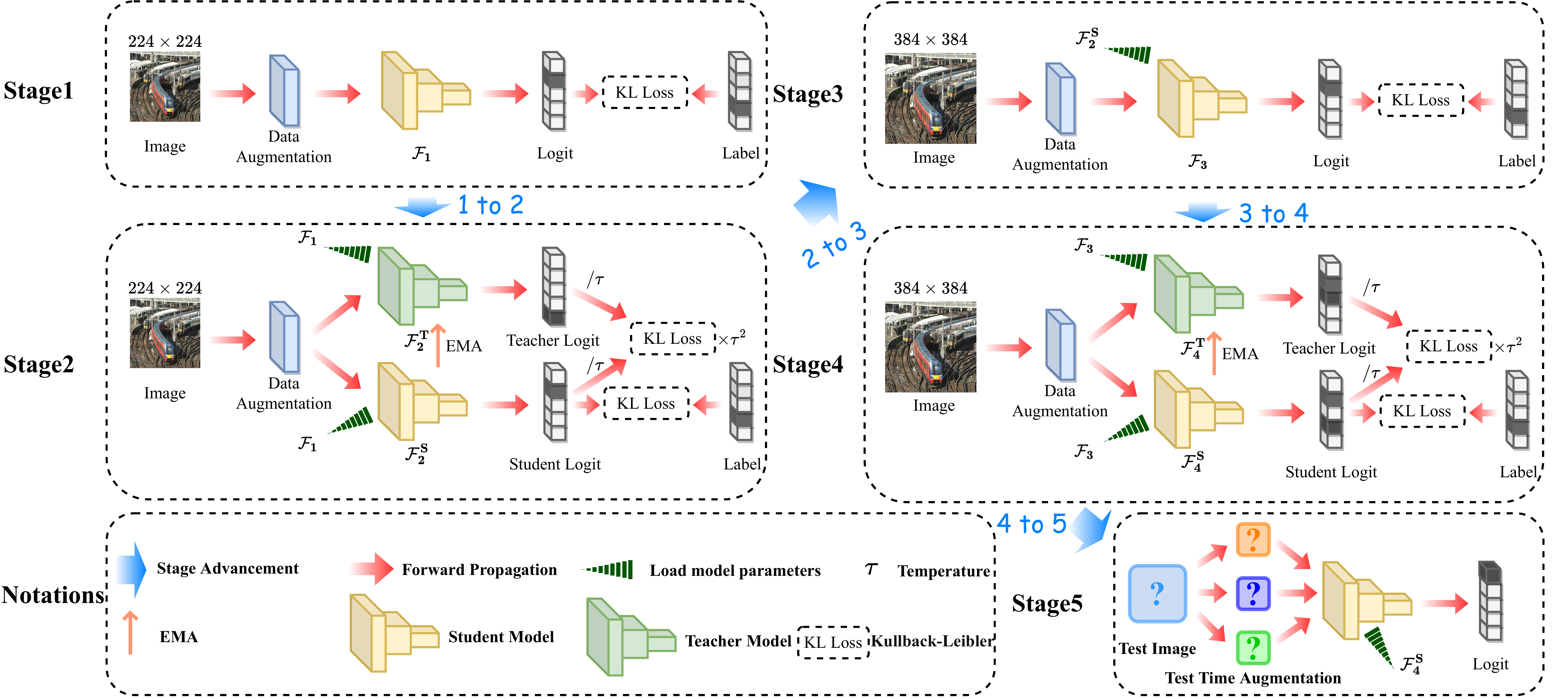}
\caption{Diagram of the general framework of the distillation-based fine-tuning paradigm. The diagram shows the overall flow from the beginning of stage1 to the end of stage 5, which includes regular pre-training at 224 resolution, supervised self-distillation pre-training at 224 resolution, fine-tuning at 384 resolution, supervised self-distillation fine-tuning at 384 resolution, and a testing stage.}
\label{fig:paradigm}
\end{figure}

\section{Method}

\subsection{Overview of Our Training Paradigm}
As shown in Fig.~\ref{fig:paradigm}, the proposed distillation-based fine-tuning paradigm is a four-step process (test phase is not included), which aims to let the model not only adapt to higher resolution, but also converge to a flatter optimum in order to have a better generalization ability. The first step is standard classification training using Cross-Entropy loss where the input resolution is 224$\times$224. The second step is training by Supervised Self-Distillation~\cite{SAM} to let our model converge to a flatter optimum in order to boost its generalization ability. This method contains a teacher model and a student model. Parameter updates for the teacher model are implemented through Exponential Moving Average (EMA). Only the student model is updated by the gradient. The loss consists of a Cross-Entropy loss same as stage 1 and a KL divergence loss whose input is the logits of the student model and the teacher model. We will discuss this in details in Section~\ref{sec:ssd}. 

The next two stages aim to adapt the model to larger resolution because using larger resolution always leads to a better result. In the third step, we change the resolution of the input image to 384$\times$384 and we still use Cross-Entropy loss in this stage. In the last step, we use Supervised Self-Distillation to improve the model's generalization ability. Just like stage two, our goal is to let the model converge to an flatter position and thus have stronger generalization performance. 

In general, we train on small resolutions first, and then fine-tune to large resolutions. Compared with training directly at large resolution from scratch, this training process is not only fast, but also the making the model works very well. In order to make the model converge to a better optimum, we will use the MESA method to train again after each training process above. Based on this idea, we propose this four stage training paradigm.

In each stage, we will reset the learning rate to the maximum. This is because the optimum in the last stage may not be optimal in the new stage, so this operation allows the model to get out of the original optimum, and reach a new and better optimum.

\subsection{Adaptive Learning Rate Scheduler}
During optimization, the learning rate is a crucial factor. A competent learning rate scheduler will allow the model to converge to flat optimality. If the learning rate is huge, although it definitely will not converge to sharp optima, but may also not converge at all. So the best way to schedule the learning rate is making the learning rate as large as possible, as long as it does not affect convergence~\cite{saxe2014exact}. So we design our learning rate scheduler as follows:
\renewcommand{\algorithmicrequire}{\textbf{Input:}}
\renewcommand{\algorithmicensure}{\textbf{Output:}}
\begin{algorithm}[h]
\small
    \caption{Procedures of ALRS}
    \label{algorithm}
    \begin{algorithmic}[1]
    \REQUIRE
     The number of current epochs $\mathcal{T}_{e}$ and the number of epochs used for learning rate warm-up $\mathcal{T}_{w}$; The current learning rate $\mathcal{T}_{\textrm{lr}}$ and the learning rate set before training $\mathcal{T}_{\textrm{tlr}}$; The loss in the current epoch $\mathcal{L}_{\textrm{n}}$ and the loss in the last epoch $\mathcal{L}_{\textrm{p}}$; The decay rate of learning rate $\alpha$; The slope threshold $\beta_1$ and the difference threshold $\beta_2$.
    \ENSURE The learning rate for the current moment optimizer update $\mathcal{T}_\sigma$.
     \IF{$\mathcal{T}_{e} = 0$}
     \STATE $\mathcal{L}_{\textrm{n}} \gets +\infty$;
     \STATE $\mathcal{L}_{\textrm{p}} \gets \mathcal{L}_{\textrm{n}}-1$;
     \ENDIF
     \IF{$\mathcal{T}_{e}\leq \mathcal{T}_{w}$}
     \STATE $\mathcal{T}_\sigma \gets \frac{\mathcal{T}_{\textrm{tlr}}\cdot\mathcal{T}_{e}}{\mathcal{T}_{w}}$;
     \ELSE
     \STATE $\delta \gets \mathcal{L}_{\textrm{p}} - \mathcal{L}_{\textrm{n}}$;
     \IF{$\left|\frac{\delta}{\mathcal{L}_{\textrm{n}}}\right|<\beta_1\ \textrm{and}\ \left|\delta\right|<\beta_2$}
     \STATE $\mathcal{T}_\sigma \gets \alpha\cdot\mathcal{T}_{\textrm{lr}}$;
     \ENDIF
     \ENDIF
    \end{algorithmic}
\end{algorithm}
We adopt a warm-up strategy because which gives about 0.4\% improvement. We set our learning rate as large as possible, to prevent our model converge to a sharp optimum. If the loss does not decrease in this epoch, we will multiply our learning rate by decay factor $\alpha$. It turns out that this learning rate schedule gives 1.5\% improvement to our model. We will conduct experiments in details in Section~\ref{sec:er}.

\subsection{Supervised Self-Distillation}
\label{sec:ssd}
This methodology comes from MESA~\cite{SAM}, which is a kind of Sharpness Aware Minimization (SAM) that allows the model converge to a flat optimum. We initialize two models. The teacher model gets the knowledge of the student model through EMA. For the student model, the loss function is defined as follows:
\begin{align}
    \mathcal{L}_{\textrm{kd}} = \frac{1}{|S|} \sum_{x_i, y_i \in S} -\mathbf{y_i}\log{f_s}(x_i)+ \textbf{\textrm{KL}}(\frac{1}{\tau}f_{S}(x_i),\frac{1}{\tau}f_{T}(x_i)),
\end{align}
where $S$ is the training dataset, $f_{S}$ is the student network, $f_{T}$ is the teacher network, and $\mathbf{y_i}$ is the one hot vector of ground truth.

But, $\mathcal{L}_{\textrm{kd}}$ does not work well if we use it from the beginning of training, so we decide to apply it after the model trained by Cross-Entropy(CE) loss. To demonstrate the correctness of this analysis, we go through Table~\ref{table:results} to show why it does not work.
\begin{table}[h]
\centering
\renewcommand\tabcolsep{18.0pt}
\caption{Classification accuracies ($\%$) of different methods.}
\label{table:results}
\resizebox{1.0\linewidth}{!}{
\begin{tabular}{c|cccc}
	\hline
	Method& CE & MESA &MESA+CE&CE+MESA\\
	\hline
	Acc. (\%)& 83 & 82 & 83 & 84.6\\
	\hline
\end{tabular}}
\vspace{-20pt}
\end{table}

This phenomenon can be comprehend from two perspectives. First, training from scratch will cause the model to converge to a suboptimal minimum point, although the landscape near this point is flat, and it will still resulting in poor model performance on both training set and test set. Besides, training by Cross-Entropy loss first and then performing MESA can be easily comprehended from distillation perspective because this is very common in distillation where training a large model first and then distillate this model.
All in all, the best way to perform MESA is first to use Cross Entropy to train a teacher model first, and then use MESA to allow the new model to converge to a flatter region, thus obtaining a better generalization ability.

\subsection{Architecture}
It turns out that wider neural network tends to converge to a flatter optimum, and have a better generalization ability~\cite{li2018visualizing}. But wider neural networks have more parameters, which is not desirable in training and inference. To address this problem, we decide to increase the width of the toeplitz matrix corresponding to the input of each layer instead of increasing the width of the neural network.

\begin{figure}[h]
\centering
\includegraphics[width=\textwidth]{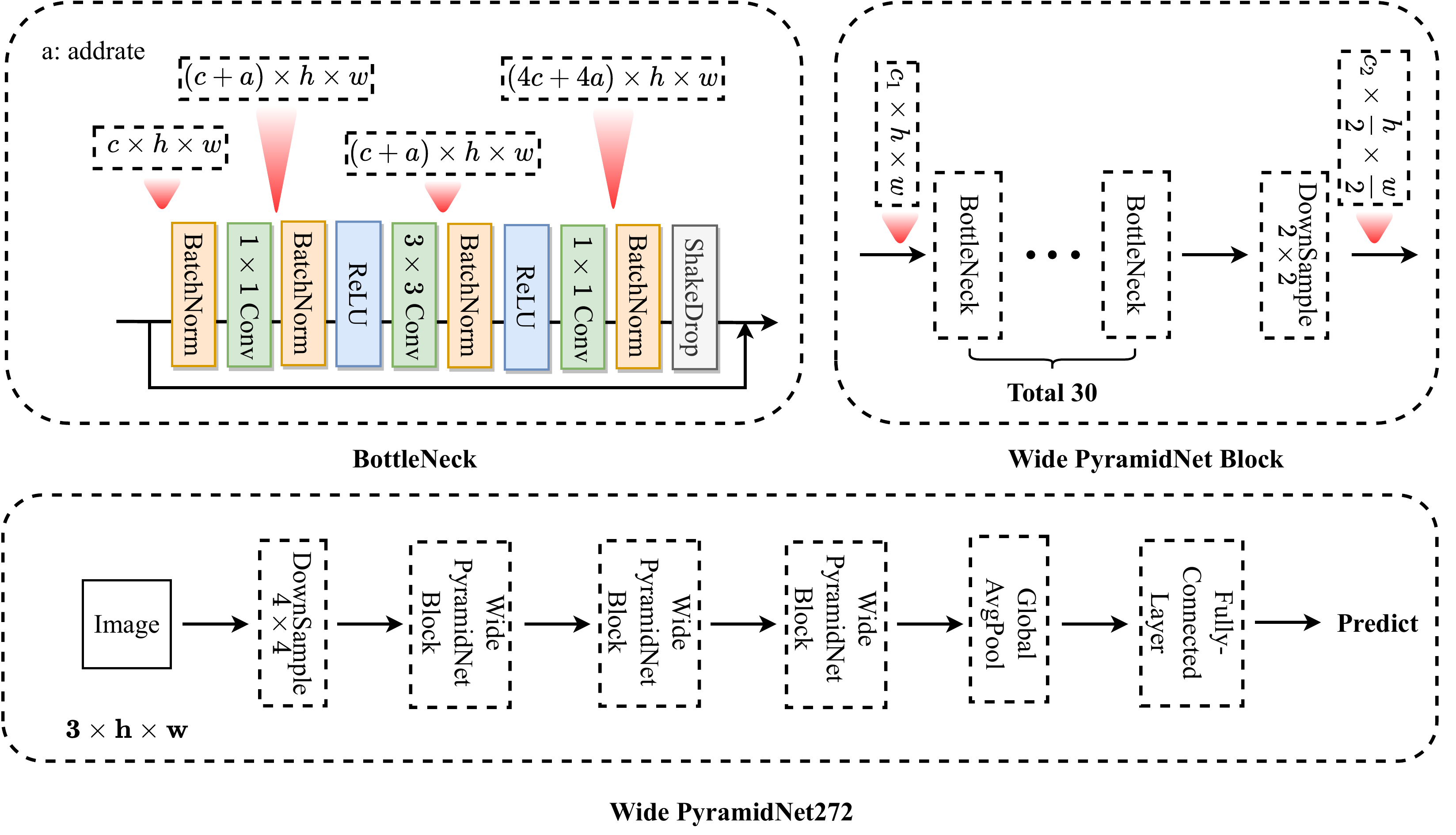}
\caption{The general model diagram of Wide PyramidNet272. Where $c_1$ and $c_2$ represent the number of channels of input and output of Wide PyramidNet Block, respectively. Compared with vanilla PyramidNet, we have increased the number of channels of the model.}
\label{fig:pyramidnet_wide}
\end{figure}

At the beginning of the model, we downsample the input to 56$\times$56 if the input size is 224$\times$224. As shown in Fig.~\ref{fig:pyramidnet_wide}, our model is wider than vanilla PyramidNet~\cite{han2017deep}. This can lead to more flat optimum and less sharp optimum~\cite{li2018visualizing}. The model has 3 stages, and each stage contains 30 blocks. For each block, we adopt bottleneck structure, and the bottleneck ratio is 4. We use ShakeDrop~\cite{yamada2019ShakeDrop} regularization at the residual connections. The structure of our bottleneck is same as the original PyramidNet, where 4 BatchNorms are interspersed next to 3 convolutions, and the kernel size of the convolution is 1$\times$1, 3$\times$3, 1$\times$1, respectively. There is only two activation function in each block, thus avoiding information loss due to activation functions.

Our neural network, like PyramidNet~\cite{han2017deep}, increases the dimension of the channel at each block. This makes our backbone very different from other Convolutional Neural Networks(CNNs), which can better defend against adversarial attacks. Since Swin Transformer~\cite{SWIN-T}, ViT~\cite{VIT} do not perform well on this dataset, we do not adopt them as backbone networks, although they are more different from CNNs.

At test time, we can apply the re-parameterization trick, so there are only three convolutions and two activation functions in each block. The inference speed will be much faster compared to the most backbone of the same size.

\subsection{Regularization}
In order to make our model have better generalization ability, we choose regularization that disturbs the backward gradient, which can make the model converge to the flatter optimum. ShakeDrop~\cite{yamada2019ShakeDrop} is a kind of mix of stochastic depth~\cite{stochasticdepth} and shake-shake~\cite{shakeshake}, where the gradient in backward pass is not same with the truth gradient, so it's hard for model to converge into sharp optimum but it's easy to converge into flat optimum. This can also lead to better generalization ability.

\begin{figure}[h]
\centering
\includegraphics[width=\textwidth]{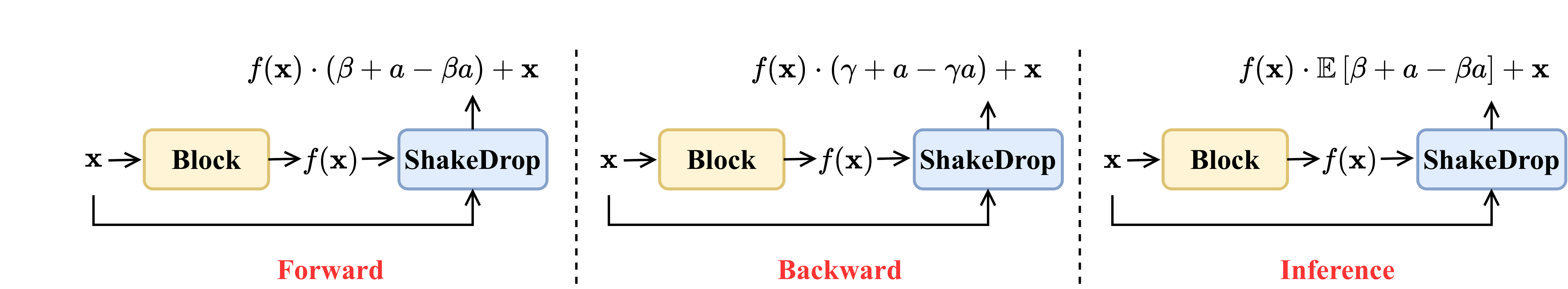}
\caption{Illustrations of ShakeDrop~\cite{yamada2019ShakeDrop}}
\label{fig:example}
\end{figure}

In every residual connection, the ShakeDrop takes the output of residual branch and block branch, and outputs a linear combination of them.
The formulation is given as
\begin{equation}
\begin{split}
    \textrm{ShakeDrop}(\mathbf{x}_\textrm{residual},\mathbf{x}_\textrm{block}) & = \mathbf{x}_\textrm{residual} + (\beta + \alpha - \beta \alpha){x}_\textrm{block} ,\\
\end{split}
\end{equation}

where $\beta$ is sampled from a Bernoulli distribution while $\alpha$ is sample from a uniform distribution.

In backward propagation, if $\alpha$ is zero, the gradient formula is denoted as
\begin{align}
    \textrm{grad}_\textrm{out} = \textrm{grad}_\textrm{in} + (\gamma+\alpha - \gamma\alpha) \textrm{grad}_\textrm{in},
\end{align}
where $\gamma$ is sampled from a uniform distribution like $\beta$, $grad_{in}$ represents the gradient of loss with respect to the ouput of this block, while the $grad_{out}$ represents the gradient of loss with respect to the input of this block.

These equations can also be understood from this perspective: $\alpha$ is used to control whether to scale the block. It turns out that the model performs best when alpha is 0.5, which indicate that there is a half probability of scaling the output of the block. $\beta$ and $\gamma$ are used to scale the output of each block and the gradient of backpropagation, respectively.

During inference, like dropout, each block only needs to be multiplied by the expectation of the coefficient
\begin{equation}
        \textrm{ShakeDrop}(x_\textrm{residual},x_\textrm{block}) = x_\textrm{residual} + \mathbb{E}\left[\beta + \alpha - \beta \alpha\right]x_\textrm{block},
\end{equation}

\section{Experiments}
This section first introduces the relevant datasets used in this paper, then describes our experimental settings, shows our series of ablation experiments and comparison experiments on NICO++~\cite{nico}, PACS~\cite{PACS} and VLCS~\cite{vlcs} datasets, and finally visualizes the loss landscape for several critical experimental results. Therefore, we effectively prove the rationality of our model design and the correctness of the loss landscape theory for this context generalization task.
\renewcommand{\arraystretch}{1.1}
\setlength{\tabcolsep}{0.05em}
\begin{table}[t]
\centering
\caption{Hyperparameter settings for different stages on the NICO++ datasets. We only show here the relatively important hyperparameter settings, the rest of the relevant details can be found in our code.}
\label{table:hyperparameter}
\begin{tabular}{l|cccc}
\hline
\multirow{2}{*}{Hyper-parameters} & \multicolumn{4}{c}{Values} \\
 & Stage1 & Stage2 & Stage3 & Stage4 \\\hline
\multirow{4}{*}{Augmentation} & AutoAugment & AutoAugment & AutoAugment & AutoAugment  \\ 
 & ColorJitter & ColorJitter & ColorJitter & ColorJitter \\
& RandomdCrop & RandomdCrop & RandomCrop & RandomdCrop \\
& CutMix & CutMix & CutMix & CutMix \\
Learning Rate & 1e-1 & 1e-1 & 1e-2 & 1e-2\\
Optimizer & SGD & SGD & SGD & SGD \\
Weight Decay & 5e-4 & 5e-4 & 5e-4 & 5e-4 \\
\multirow{2}{*}{\begin{tabular}[l]{@{}l@{}}  Temperature \\ (knowledge distillation)\end{tabular}}& \multirow{2}{*}{-} & \multirow{2}{*}{5} &  \multirow{2}{*}{-} & \multirow{2}{*}{5} \\
&&&&\\
LR Decay Rate $\alpha$ & 0.9 & 0.9 & 0.8 & 0.8 \\
Slope Threshold $\beta_1$ & 0.2 & 0.2 & 0.2 & 0.2\\
Difference Threshold $\beta_2$ & 0.2 & 0.2 & 0.2 & 0.2\\
Minimum Learning Rate & 1e-4 & 1e-4 & 1e-5 & 1e-5 \\
Batch Size & 48 & 48 & 16 & 16 \\
Image Size& 224$\times$224 & 224$\times$224 & 384$\times$384 & 384$\times$384\\
\multirow{2}{*}{\begin{tabular}[l]{@{}l@{}} Number of Epochs in  \\ Test Time Augmentation \end{tabular}}& \multirow{2}{*}{80} & \multirow{2}{*}{80} & \multirow{2}{*}{80} &  \multirow{2}{*}{80} \\
&&&&\\
Resource Allocation & Two RTX 3090 & Two RTX 3090 & Two RTX 3090 & Two RTX 3090 \\
\hline
\end{tabular}
\end{table} 

\subsection{Datasets}
We validate our proposed algorithm on three domain generalization datasets named NICO++, PACS, and VLCS. We only compare the relevant state-of-the-art (SOTA) algorithms in recent years. Also, since we use PyramidNet272 as the backbone, we additionally append the comparison results of ResNet18~\cite{ResNet}.

\noindent\textbf{NICO++} is comprised of natural photographs from 10 domains, of which 8 are considered as sources and 2 as targets. In accordance with~\cite{stable_net}, we randomly divided the data into 90\% for training and \% for validation, reporting metrics on the left domains for the best-validated model.

\noindent\textbf{PACS} contains a total of 9991 RGB three-channel images, each of which has a resolution of 227$\times$227. The dataset is divided into seven categories and four domains (i.e., \textbf{P}hoto, \textbf{A}rt, \textbf{C}artoon, and \textbf{S}ketch). Following EntropyReg \cite{EntropyReg}, we utilize the split file for training, validating, and testing. The training and validation sets are comprised of data from the source domains, whereas the test set is comprised of samples from the target domain. We select classifiers in accordance with the validation metric.

\noindent\textbf{VLCS} is a dataset for evaluating the domain generalization ability of a model, which contains a total of 5 categories and 4 domains (i.e., Pascal \textbf{V}OC2007 \cite{voc}, \textbf{L}abelMe \cite{labelme}, \textbf{C}altech \cite{caltech}, and \textbf{S}UN09 \cite{sun}). We fixedly divide the source domains into training and validation sets with a ratio of 9 to 1.

\subsection{Settings}
Our proposed DFP is composed of four different stages, so we have different hyper-parameter settings in different stages. As shown in Table~\ref{table:hyperparameter}, the first two stages belong to the low-resolution pre-training stage, and the optimizer is SGD. The last two stages belong to the high-resolution fine-tuning stage, and the optimizer is AdamW. In addition, there is no hyper-parameter such as the so-called total epoch in our model training since the novel ALRS is used to schedule the learning rate. In other words, when the learning rate saved by the optimizer is less than the artificially set minimum learning rate, the model training is terminated at that stage.

\subsection{Domain generalization results on NICO++ dataset}
\label{sec:er}

This subsection presents a series of ablation experiments and comparison experiments carried out on the NICO++ dataset, including the effect of different initial learning rates on model robustness, the impact of ALRS using different lr decay rates, a comparison of test accuracy at DFP's different stages, the effect of different backbones on test performance, and the negative effects of domain-invariant representation learning methods.

\begin{wraptable}{l}{0.5\linewidth}
\centering
\vspace{-30pt}
\renewcommand\tabcolsep{6.0pt}
\caption{\textbf{Acc.} Top-1 Test Accuracy [\%]. The table shows the impact on the final performance of the model when we use different initial learning rates based on a fixed backbone named PyramidNet272.}
\label{table:learning_rate_results}
\resizebox{1.0\linewidth}{!}{%
\begin{tabular}{c|ccccc}
	\hline
	Learning Rate& 0.5 & 0.1 &0.05&0.01&0.005\\
	\hline
	Acc.& 83.0 & 83.1 & 83.1 & 81.6 & 80.5\\
	\hline
\end{tabular}}
\vspace{-20pt}
\end{wraptable}
\paragraph{Learning Rate.} We first conducted experiments comparing the final performance of the model with different initial learning rates. It turns out that if the maximum learning rate was greater than 0.05, the models performed well and had almost the same accuracy. However, if the learning rate is less than 0.05, a dramatic drop in model performance occurs. Therefore, in all experiments for this challenge, we want to ensure that the learning rate is large enough to guarantee the final model performance. It is worth noting that the experimental results are well understood, as a small learning rate for a simple optimizer such as SGD can cause the model to converge sharply to a locally suboptimal solution.
\begin{wraptable}{l}{0.5\textwidth}
\centering
\vspace{-30pt}
\caption{\textbf{Epoch.} The number of epochs required to train the model to the set minimum learning rate 1e-4. The table shows the final performance of the model training for different values of lr decay rate and the number of epochs required. The rightmost column represents the training results obtained applying the cosine learning rate scheduler.}
\label{table:lr_decay_rate}
\resizebox{1.0\linewidth}{!}{%
\begin{tabular}{l|cccc}
\hline
LR Decay Rate & 0.9 & 0.8 & 0.7 & - \\ \hline
Top-1 Test Accuracy [\%] & 83.1 & 82.8 & 81.9 & 81.5\\
Epoch & 134 & 112 & 72 & 200 \\ \hline
\end{tabular}}
\vspace{-50pt}
\end{wraptable} 
\paragraph{LR Decay Rate.} Our proposed ALRS is worthy of further study, especially its hyper-parameter called lr decay rate. Intuitively, lr decay rate represents the degree of decay in the learning rate during the training of the model. If lr decay rate is small, then the number of epochs required to train the model is small, and conversely if lr decay rate is large, then the number of epochs required to train the model is large. Referring to Table~\ref{table:lr_decay_rate}, we can see that the performance of ALRS crushes that of the cosine learning rate scheduler. Meanwhile, we can conclude that the larger the lr decay rate, the greater the computational cost spent in model training, and the better the final generalization ability. Therefore, setting a reasonable lr decay rate has profound implications for model training.

\paragraph{Different Stages in DFP.} DFP is a paradigm that combines supervised self-distillation and fine-tuning, where supervised self-distillation and fine-tuning are interspersed throughout the four stages of DFP to maximize the generalization capability of the model. The final results are presented in Table~\ref{table:dfp}, \begin{wraptable}{l}{0.5\textwidth}
\centering
\renewcommand\tabcolsep{6.0pt}
\vspace{-20pt}
\caption{\textbf{Acc.} Top-1 Test Accuracy [\%]. From the table, it can be concluded that the model's accuracy is steadily increasing from stage1 to stage4.}
\label{table:dfp}
\resizebox{1.0\linewidth}{!}{%
\begin{tabular}{cccccc}
\hline
 & TTA & \multicolumn{4}{c}{No. of stage} \\ 
 & & No.1& No.2 & No.3 & No.4 \\ \hline
\multirow{2}{*}{Acc.}& $\times$  & 83.1 & 84.6 & 86.8 & 87.1\\ 
& \checkmark & 84.7 & 86.3 & 87.5 & 87.8\\ \hline
\end{tabular}}
\vspace{-35pt}
\end{wraptable} 
which shows that the performance of the different stages is related as follows: stage1$<$stage2$<$stage3$<$stage4. This result also explicitly indicates that each stage in DFP is necessary and not redundant.

\paragraph{Backbone.} Backbone, as the feature extraction module in this context generalization task, is the main factor affecting the model's final generalization ability. Therefore, we conducted ablation experiments on Backbone with the expectation of selecting a suitable feature extractor and performing well in domain generalization. The experimental results are shown in Table~\ref{table:backbone}, where we can easily find that PyramidNet272 performs the best. We designed Wide PyramidNet272 inspired by the loss landscape, which has an accuracy of 83.1 on the test set, an improvement of 1.6 points compared to the vanilla PyramidNet272.
\begin{table}[H]
\centering
\caption{\textbf{Acc.} Top-1 Test Accuracy [$\%$]. Even with inductive bias, the table shows that Swin Transformer performs badly on the NICO++ dataset. So, when making our choice of Backbone, the ViT series was not taken into account. In addition, PyramidNet272 also has the best performance, which is an essential reason we select it as the Backbone.}
\label{table:backbone}
\resizebox{1.\linewidth}{!}{
\begin{tabular}{c|cccc}
\hline
Backbones & DenseNet190~\cite{huang2017densely} & DenseNet201~\cite{huang2017densely} & PyramidNet272~\cite{han2017deep} & PyramidNet101~\cite{han2017deep} \\
Acc. & 77.3 & 73.6 & 81.5 & 78.1 \\\hline
Backbones &  WRN28-10~\cite{Zagoruyko2016WRN} & RegNet\_x\_32gf~\cite{radosavovic2020designing} & Swin-B~\cite{SWIN-T} & - \\
Acc. & 72.4 & 77.6 & 46.2 & - \\
\hline
\end{tabular}}
\end{table} 

\begin{wraptable}{l}{0.5\textwidth}
\centering
\vspace{-30pt}
\caption{\textbf{Acc.} Top-1 Test Accuracy [$\%$]. One obvious conclusion that can be drawn from the table is that all domain-invariant methods fail on the NICO++ dataset.}
\label{table:invariant}
\resizebox{1.0\linewidth}{!}{%
\begin{tabular}{cccc}
\hline
Methods& Baseline& EFDM~\cite{zhang2022exact}& DANN~\cite{ganin2016domain} \\
Acc. & 83.1 & 82.7& 79.2 \\\hline
\end{tabular}}
\vspace{-35pt}
\end{wraptable}  
\paragraph{Domain-invariant Representation Learning.} As shown in Table~\ref{table:invariant}, we experimentally demonstrate that all domain invariant representation learning methods fail in the NICO++ dataset. Therefore, this forces us to explore other theories (i.e., loss landscape perspective) to perform domain generalization.
\begin{table}[th]
\renewcommand\tabcolsep{10.0pt}
\renewcommand{\arraystretch}{1.1}
\begin{center}
\caption{Leave-one-domain-out classification accuracy(\%) on PACS. Best performances are highlighted in bold.} 
\label{tab:pacs}
\resizebox{1.0\textwidth}{0.25\textheight}{
\begin{tabular}{l|c|cccc|c}
\toprule
& D\_ID & P & A & C & S & Avg. \\
\midrule
\multicolumn{7}{l}{\textit{AlexNet}}\\
\midrule
DSN~\cite{dsn} &\cmark & 83.30 & 61.10 & 66.50 & 58.60 & 67.40 \\
Fusion~\cite{fusion} &\cmark & 90.20 & 64.10 & 66.80 & 60.10 & 70.30\\
MetaReg~\cite{MetaReg} &\cmark & 87.40 & 63.50 & 69.50 & 59.10 & 69.90  \\
Epi-FCR~\cite{EPI-FCR} &\cmark & 86.10 & 64.70 & 72.30 & 65.00 & 72.00\\
MASF~\cite{MASF} &\cmark & 90.68 & 70.35 & 72.46 & 67.33 & 75.21\\
DMG~\cite{DMG} &\cmark & 87.31 & 64.65 & 69.88 & 71.42 & 73.32\\
\hline
HEX~\cite{HEX} &\xmark & 87.90 & 66.80 & 69.70 & 56.20 & 70.20\\
PAR~\cite{PAR} &\xmark & 89.60 & 66.30 & 66.30 & 64.10 & 72.08\\
JiGen~\cite{JIGSAW} &\xmark &89.00& 67.63&71.71&65.18&73.38\\
ADA~\cite{ADA} &\xmark & 85.10& 64.30& 69.80& 60.40 & 69.90\\
MEADA~\cite{MEADA} &\xmark & 88.60 & 67.10 & 69.90& 63.00 & 72.20\\
MMLD~\cite{mmld} &\xmark & 88.98 & 69.27 & 72.83 & 66.44 & 74.38\\
\midrule
\multicolumn{7}{l}{\textit{ResNet18}}\\
\midrule
Epi-FCR &\cmark& 93.90 & 82.10 & 77.00 & 73.00 & 81.50 \\
MASF &\cmark& 94.99 & 80.29 & 77.17 & 71.68 & 81.03 \\
DMG &\cmark& 93.55 & 76.90 & 80.38 & 75.21 & 81.46 \\
\hline
Jigen &\xmark& 96.03 & 79.42 & 75.25 & 71.35 & 80.51 \\
ADA &\xmark& 95.61 & 78.32 &77.65 & 74.21 &81.44 \\
MEADA &\xmark& 95.57 & 78.61 & 78.65 & 75.59 & 82.10\\
MMLD &\xmark& 96.09 & 81.28 & 77.16 & 72.29 & 81.83\\
\hline
Ours &\xmark& 96.95 & 78.54 & 75.97 & 83.35 & 83.70 \\
\midrule
\multicolumn{7}{l}{\textit{PyramidNet272}}\\
\midrule
Ours &\xmark& \textbf{98.50} & \textbf{87.21}  & \textbf{82.76} & \textbf{88.24} & \textbf{89.18} \\
\bottomrule
\end{tabular}}
\end{center}

\end{table} 
\setlength{\tabcolsep}{10pt}
\begin{table}[bth!]
\begin{center}
\caption{Leave-one-domain-out results on VLCS dataset. Our proposed ALRS+DFP achieves optimal results on two target domains (i.e., LabelMe and SUN09) and outperforms all related methods in terms of average metrics.}
\label{tab:vlcs}
\resizebox{1.0\textwidth}{.13\textheight}{
\begin{tabular}{l|cccc|c}
\hline
Method      & VOC & LabelMe & Caltech & SUN09 & Avg.  \\ \hline
DBADG~\cite{PACS}       & 69.99      & 63.49   & 93.64   & 61.32 & 72.11 \\
ResNet-18   & 67.48      & 61.81   & 91.86   & 68.77 & 72.48 \\
JiGen       & 70.62      & 60.90   & 96.93   & 64.30 & 73.19 \\
MMLD        & 71.96      & 58.77   & 96.66   & 68.13 & 73.88 \\
CIDDG~\cite{CIDDG}       & 73.00      & 58.30   & 97.02   & 68.89 & 74.30 \\
EntropyReg~\cite{EntropyReg}  & 73.24      & 58.26   & 96.92   & 69.10 & 74.38 \\
GCPL~\cite{gcpl}        & 67.01      & 64.84   & 96.23   & 69.43 & 74.38 \\
RSC~\cite{RSC}         & \textbf{73.81}      & 62.51   & 96.21   & 72.10 & 76.16 \\
StableNet~\cite{stable_net}   & 73.59      & 65.36   & 96.67   & 74.97 & 77.65 \\
PoER~\cite{PoER} & 69.96      & 66.41  & \textbf{98.11}   & 72.04 & 76.63      \\ \hline
Ours & 64.23     & \textbf{74.13}  & 97.24   & \textbf{76.45} & \textbf{78.01}   \\ \hline
\end{tabular}}
\end{center}
\vspace{-12pt}
\end{table}

\subsection{Leave-one-domain-out results on PACS dataset}

The results of our method compared with other related approaches on the PACS dataset are shown in Table~\ref{tab:pacs}. For a fair comparison, we performed additional experiments on ResNet18. It is worth noting that since the maximum PACS resolution is 227$\times$227, we discarded stages 3 and 4 in the DFP paradigm, but even so, we outperformed all relevant methods on the ResNet18 benchmark. In addition, the generalization ability of the model is greatly improved when we use PyramidNet272 as the backbone.

\subsection{Leave-one-domain-out results on VLCS dataset.} On the VLCS multi-source domain generalization dataset, we divide the source domain into 70\% training data and 30\% validation data for the experiments as we go along. The experimental results in Table~\ref{tab:vlcs} show that our method beats all relevant previous techniques and achieves optimal results when only applying ResNet18 as the backbone. Therefore, getting a flatter convergence region by DFP is able to improve the model generalization ability.

\subsection{Visualizations}
\begin{figure}[bhtp!]
\centering
\includegraphics[width=0.6\textwidth]{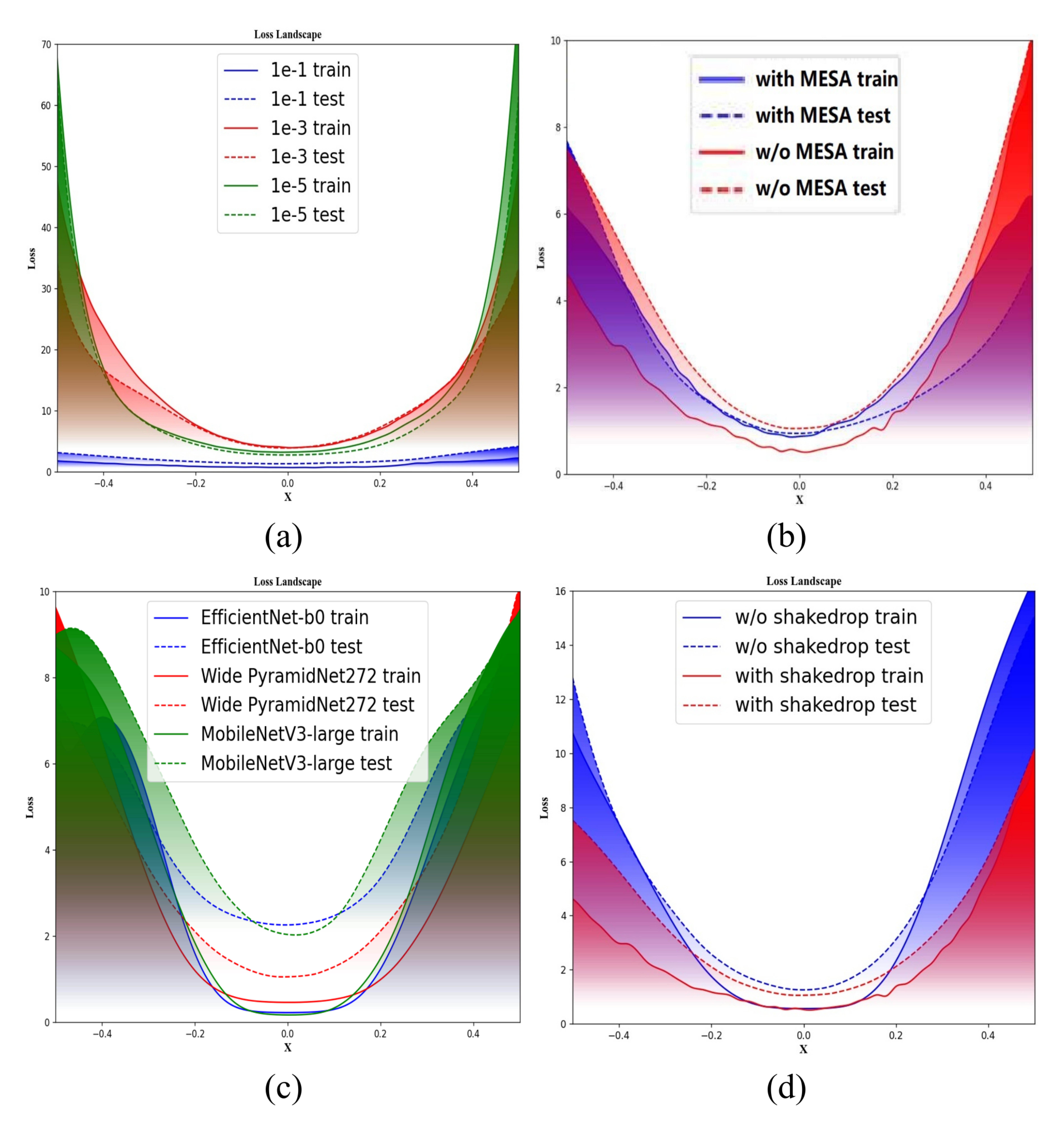}
\caption{Visualization of the loss landscape for a series of factor comparisons. \textbf{(a)} different learning rates; \textbf{(b)} whether to use knowledge distillation; \textbf{(c)} different backbone; \textbf{(d)} whether to use ShakeDrop for regularization constraint.}
\label{fig:losslandscape}
\end{figure}

\paragraph{Learning Rate.} As shown in Fig.~\ref{fig:losslandscape}(a), we draw the loss landscape under different learning rates. We can observe that the landscape near the optimum is flatter when the learning rate is relatively large. So the model can't converge to a sharp optimum. The model may converge to a sharp optimum when a relatively low learning rate. Therefore, from our point of view, we out to increase the maximum learning rate as much as possible for image classification tasks.

\paragraph{Supervised Self-Distillation.} As shown in Fig.~\ref{fig:losslandscape}(b), we compared the loss landscape of models trained by distillation and those without distillation. It can be observed that the loss landscape of the model trained with distillation is flatter than the one without MESA. This has been theoretically and empirically proved by~\cite{SAM}.

\paragraph{Backbone.} As shown in Fig.~\ref{fig:losslandscape}(c), we compare the lost landscape of our backbone and other backbones. Because our backbone increases the width of the Toeplitz matrix, the loss landscape near the optimum that our model converges on is flatter. This idea has been verified in ~\cite{li2018visualizing} and ~\cite{wideresnet}.

\paragraph{Regularization.} As shown in Fig.~\ref{fig:losslandscape}(d), the model with ShakeDrop converges more easily to the flat region than the model without ShakeDrop. This is because ShakeDrop perturbs the backward gradient, making it difficult for the model to converge stably to the sharp optimum.

\section{Conclusion}
Overall, we bootstrap the generalization ability of the deep learning model from a landscape perspective in four dimensions, namely backbone, regularization, training paradigm, and learning rate, with the goal of allowing the model to converge to a flatter optimum. On the basis of these analyses, we propose DFP with ALRS and validate and attain the best performance in a variety of domain generalization datasets, including PACS and VLCS datasets, using a variety of approaches. In addition, we have conducted numerous ablations and visualizations on NICO++ dataset to validate the accuracy and viability of our approaches. In future research, we will broadly apply our methods to training deep learning models, enabling the models to acquire greater generalizability with fewer data.

\subsubsection{\ackname} 
This work was supported in part by the Natural Science Foundation of China (NSFC) under Grants No. 62072041.

\bibliographystyle{splncs04}
\bibliography{egbib}
\end{document}